\documentclass[11pt]{article}

\usepackage{lineno}
\modulolinenumbers[5]

\usepackage[pdftex]{graphicx}
\usepackage[pdftex]{color}

\usepackage[square]{natbib}
\bibpunct[:]{(}{)}{;}{a}{}{:}
\usepackage{amssymb}
\usepackage{color}

\newcommand{\msk}{\hspace*{3mm}}

\usepackage{graphicx}

\newcommand{\eqref}[1]{Formula~ \ref{#1}}
\newcommand{\secref}[1]{\S \ref{#1}}
\newcommand{\figref}[1]{Figure~\ref{#1}}

\title{
  Long-Range Correlation Underlying Childhood Language \\
  and Generative Models
}

\date{\today}
\author{Kumiko Tanaka-Ishii \\
  Research Center for Advanced Science and Technology, University of Tokyo\\
4-6-1 Komaba, Meguro-ku, Tokyo 153-8904, Japan. \\  
  {\tt kumiko@cl.cast.u-tokyo.ac.jp}
}

\begin{document}

\maketitle

\abstract{ Long-range correlation, a property of time series
  exhibiting long-term memory, is mainly studied in the statistical
  physics domain and has been reported to exist in natural
  language. Using a state-of-the-art method for such analysis,
  long-range correlation is first shown to occur in long CHILDES data
  sets. To understand why, Bayesian generative models of language,
  originally  proposed in the cognitive scientific domain,  are
  investigated. Among representative models, the Simon model was found
  to exhibit surprisingly good long-range correlation, but {\em not}
  the Pitman-Yor model. Since the Simon model is known not to
  correctly reflect the vocabulary growth of natural language, a
  simple new model is devised as a conjunct of the Simon and
  Pitman-Yor models, such that long-range correlation holds with a
  correct vocabulary growth rate. The investigation overall suggests
  that uniform sampling is one cause of long-range correlation and
  could thus have a relation with actual linguistic processes.

}

\section{Introduction}
\label{sec:introduction}
State-of-the-art Bayesian mathematical models of language include the
Simon and Pitman-Yor models and their extensions \citep{pitman}
\citep{goldwater_ml} \citep{lee14} \citep{chater08}. These models have
been not only successful in modeling language from a cognitive
perspective but also applicable in natural language engineering
\citep{teh06}. They have been adopted primarily because the
rank-frequency distribution of words in natural language follows a
power law. Advances in studies on the statistical nature of language
have revealed other characteristics besides Zipf's law. For example,
Heaps' law describes how the growth of vocabulary forms a power law
with respect to the total size \citep{Guiraud1954} \citep{Herdan1964}
\citep{heaps}; the Pitman-Yor model follows this principle well.

In this paper, another power law underlying the autocorrelation
function of natural language is considered. Called long-range
correlation, it captures a qualitatively different characteristic of
language. As described in detail in the following section, long-range
correlation is a property of time series that has mainly been studied
in the statistical physics domain for application to natural and
financial phenomena, including natural language. When a text has
long-range correlation, there exists a (yet unknown) structure
underlying the arrangements of words. One rough, intuitive way to
understand this is by the tendency of rare words to cluster. The
phenomenon is actually more complex, however, because it has been
reported to occur at a long-range scale. Since the methods used to
investigate this phenomenon measure the similarity between two long
subsequences within a sequence, long-range correlation suggests some
underlying self similarity. In other words, it is not only the case
that rare words cluster, but more precisely, that words at all
different rarity levels tend to cluster.

Verification of the universality of long-range correlation in language
is an ongoing topic of study and has been reported across domains. In
linguistics, it has been shown through hand counting how rare words
cluster in the {\em Iliad} \citep{boas04}.  Computational methods from
the statistical physics domain have given multiple indications of the
existence of long-range memory in literary texts \citep{Ebeling1994}
\citep{Altmann2012} \citep{plosone16}. Moreover, long-range
correlation has been reported to occur across multiple texts
\citep{serrano09} and also in news chats \citep{Altmann2009}.  In
recent years, using the methods proposed in the statistical physics
domain, analysis of long-range correlation is reported: For example,
\cite{seron2014} shows how social interaction is long-range correlated
by using detrended fluctuation analysis, and \cite{ruiz2014} shows how
skilled piano play also is long-range correlated and how it is related
to auditory feedback.

In this article, first, long-range correlation for long sets of
CHILDES data is reported. The fact that power-law behavior exists in
early childhood language is surprising, since children's linguistic
utterances seem undeveloped, lacking vocabulary and proper structure,
and full of grammatical errors. Given the power law indicated by the
autocorrelation function, there must be an innate mechanism for the
human language faculty.

To explore the source of this mechanism, the article investigates how
this autocorrelative nature is present in Bayesian models, one
state-of the art models, originating in psychology.  The sequences
generated by a Pitman-Yor model \citep{pitman} are not long-range
correlated, which raises a question of the validity of Pitman-Yor
models in scientific language studies. In contrast, the Simon model
\citep{simon55}, the simplest model commonly adopted in complex
systems studies, has strong long-range correlation. Given how the
Simon model works, this suggests that one cause of 
autocorrelative nature lies in uniform sampling from the past sequence along with
introduction of new words from time to time. Since the Simon model has
a drawback with respect to vocabulary growth, a simple conjunct model
is defined so as to produce both long-range correlation and correct
vocabulary growth. In conclusion, the article discusses the relation
between uniform sampling and linguistic procedures.

\section{Quantification of Long-Range Correlation}
\label{sec:law}
\label{sec:acf}

\begin{figure} [t]
\centering
\includegraphics[width=\textwidth]{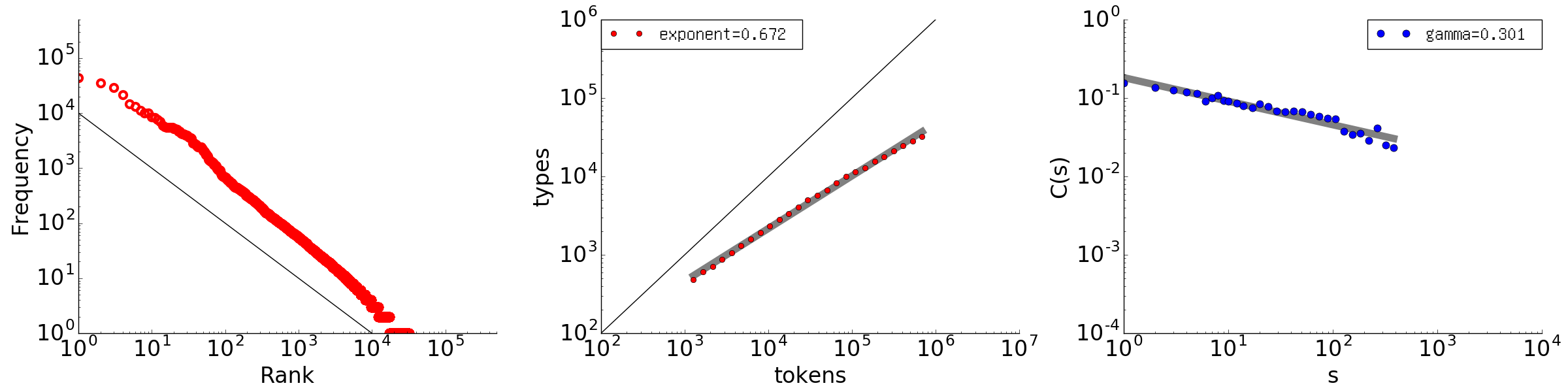}
\caption{Log-log plots of the rank-frequency distribution, type-token
  relation, and autocorrelation function for {\em Les Mis\'{e}rables}
  (by V. Hugo (French), number of words $691407$). {\bf Left:}
  Rank-frequency distribution, where the x-axis indicates the rank,
  and the y-axis indicates the frequency. The red points represent the
  actual data, and the black line indicates a slope of $\xi=1.0$. {\bf
    Middle:} Type-token relation, where the x-axis indicates the text
  size in words, and the y-axis indicates the vocabulary size. The set
  of red points represents the actual data, along with its fit line in
  light gray, and the black line indicates a slope of $\zeta=1.0$. The
  fitted exponent is shown in the upper left corner. {\bf Right:}
  Autocorrelation function applied to intervals, where the x-axis
  indicates the offset $s$, and the y-axis indicates the value of the
  autocorrelation function $C(s)$ for an interval sequence consisting
  of one-sixteenth of all the words.  The blue points represent the
  actual data, the thick gray line is the fitted power law, and the
  slope is shown in the upper right corner.}
\label{fig:lesmiserables}
\end{figure}

The focus of this paper is the power law observed for the
autocorrelation function when applied to natural language. As an
example, the rightmost graph in \figref{fig:lesmiserables} shows the
autocorrelation function applied to the text of {\em Les
  Mis\'{e}rables}. The points are aligned linearly in a log-log plot,
so they follow a power law. The correlation is {\em long}, in contrast
to short-range correlation, in which the points drop much earlier in
an exponential way.

There is a history of nearly 25 years of great effort to quantify this
long-range correlation underlying text. Since all the existing
analysis methods for quantifying long-range memory---i.e., the
autocorrelation function that is defined and used later in this
section, fluctuation analysis \citep{Kantelhardt2001}
\citep{Kantelhardt2002}, and the older Hurst method
\citep{Hurst}---apply {\em only} to numerical data, much effort has
focused on the question of how best to apply these methods to
linguistic (thus, non-numerical) sequences. Previous studies applied
one of these methods to a binary sequence based on a certain target
word \citep{Ebeling1994} \citep{Altmann2012}, a word sequence
transformed into corresponding frequency ranks \citep{Montemurro2014},
and so on. State-of-the-art approaches use the concept of intervals
\citep{Altmann2009} \citep{plosone16}, with which a numerical sequence
is derived naturally from a linguistic sequence. Note that this
transformation into an interval sequence is not arbitrary compared with
other transformations, such as the one into a rank sequence. An
approach using only interval sequences, however, suffers from the
low-frequency problem of rare words, and clear properties cannot be
quantified even if they exist. Here, instead, the analysis uses the
method proposed in \citep{plosone16}, which combines interval analysis
and extreme value analysis and has been rigorously established,
applied, and validated in the statistical physics domain
\citep{Lennartz2009}. A self-contained summary of the analysis scheme
is provided here, and a detailed argument for the method is found in
\citep{plosone16}.

The method basically uses the autocorrelation function to quantify the
long-range correlation. Given a numerical sequence
$R=r_1,r_2,\ldots,r_{M}$, of length $M$, let the mean and standard
deviation be $\mu$ and $\sigma$, respectively. Consider the following
autocorrelation function:
\begin{equation}
 C(s)= 
\frac{1}{(M-s)\sigma^2}\sum_{i=1}^{M-s}(r_i-\mu)(r_{i+s}-\mu). \label{acf}
\end{equation}
This is a fundamental function to measure the correlation, the
similarity of two subsequences of $s$ distance apart: it calculates
the statistical co-variance between the original sequence and a
sub-sequence starting from the $s$th offset element, standardized by
the original variance of $\sigma^2$. For every $s$, the value ranges
between $-1.0$ and $1.0$, with $C(0) = 1.0$ by definition. For a
simple random sequence, such as a random binary sequence, the function
gives small values fluctuating around zero for any $s$, since the
sequence has no correlation with itself. The sequence is judged to be
long-range correlated when $C(s)$ decays by a power law, as denoted in
the following:
\begin{equation}
C(s)=C(1)s^{-\gamma}, s>0. 
\end{equation}
The particularity of the autocorrelation lies in its long-range
nature: two subsequences existing in a sequence remain similar even if
$s$ becomes fairly large. Short-term memory, which gives the
correlation only for small $s$, shows how the target relies only on
local arrangements, in a Markovian way. In contrast, the long-range
correlation is considered important precisely because such correlation
lasts long. For a natural language sequence, too, $C(s)$ can be
calculated, and whether it exhibits power-law decay can be
verified. The essential problem lies in the fact that a language
sequence is not numerical and thus must be transformed into some
numerical sequence.

\begin{figure} [t]
\centering
\includegraphics[width=10cm]{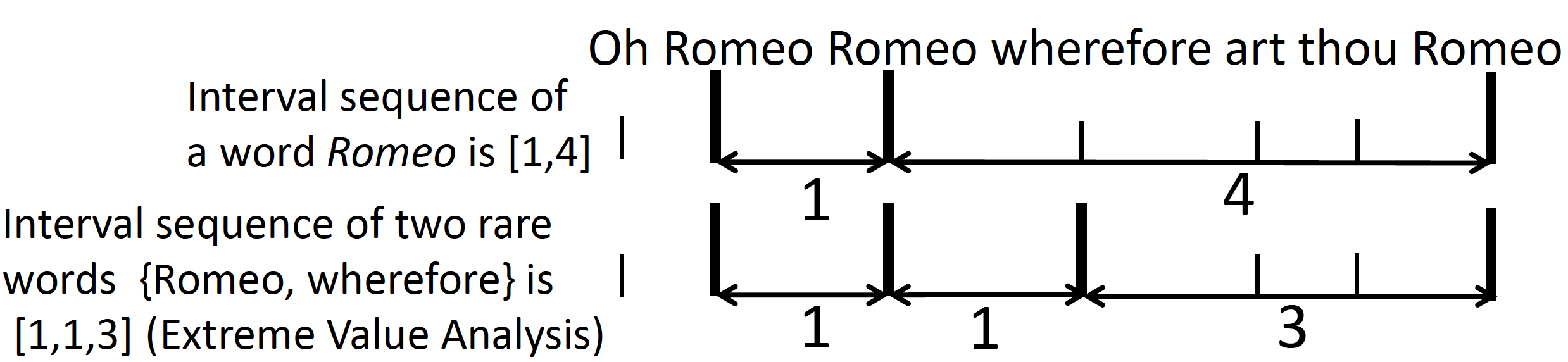}
\caption{A toy example of interval analysis and extreme value analysis.}
\label{illustration}
\end{figure}

The method of \citep{plosone16} transforms a word sequence into a
numerical sequence by using intervals of rare words. The following
example demonstrates how this is done. Consider the target {\em Romeo}
in the sequence ``Oh Romeo Romeo wherefore art thou Romeo,'' shown in
\figref{illustration}. {\em Romeo}, indicated by the thick vertical
bar, has a one-word interval between its first and second occurrences,
and the third {\em Romeo} occurs as the fourth word after the second
{\em Romeo}.  This gives the numerical sequence $[$1, 4$]$ for this
clause and the target word {\em Romeo}. The target does not have to be
one word but could be any element in a set of words. Suppose that the
target consisted of two words, the two rarest words in this clause:
{\em Romeo}, and {\em wherefore}. Then, the interval sequence would be
$[$1,1,3$]$, since {\em wherefore} occurs right after the second {\em
  Romeo}, and the third {\em Romeo} occurs as the third word after
{\em wherefore}. Since rare words occur in such small numbers,
consideration of multiple rare words serve to quantify their
behavior as an accumulated tendency.

\begin{figure} [t]
\centering
\includegraphics[width=\textwidth]{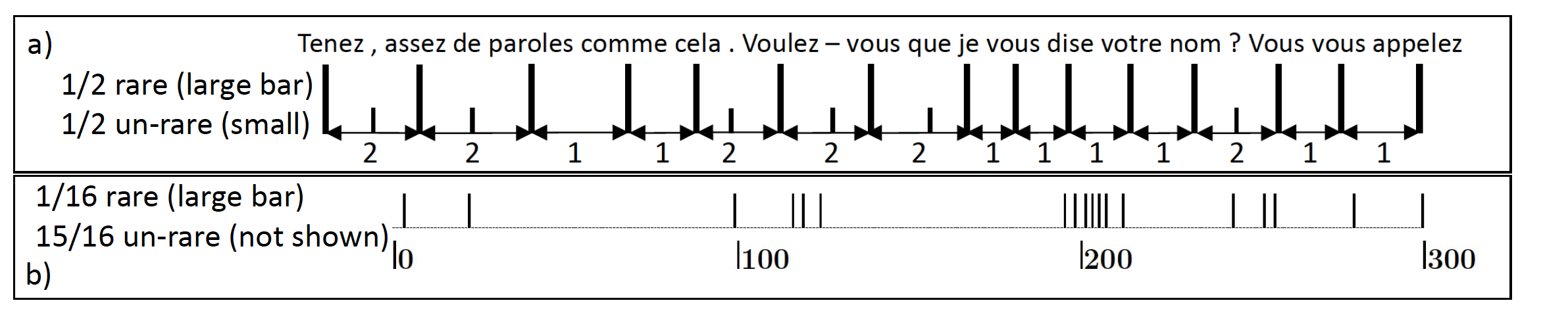}
\caption{Illustrations of (a) the procedure to acquire an interval
  sequence and (b) how rare words are clustered in a part of {\em Les
    Mis\'{e}rables}.}
\label{fig:explain}
\end{figure}

\figref{fig:explain} illustrates the analysis scheme for a longer
sequence. The upper portion (a) shows an example from {\em Les
  Mis\'{e}rables} in which half the words in the text are considered
rare (large bars), and the other half, common (small bars). By using
the large bars, the text portion is transformed into an interval
sequence shown at the bottom as $[$2,2,1,$\ldots$$]$, similarly to the
Romeo example in \figref{fig:explain}. In the bottom portion (b), one
sixteenth of all the words, instead of half, are considered rare. The
locations of only the large bars are shown for a passage of 300 words
starting from the 31096th word in {\em Les Mis\'{e}rables}: the bars
appear in a clustered manner.

As a summary, the overall procedure is described as follows. Given a
numerical sequence of length $M$, the interval length for one $N$th of
(rare) words would be $M_N \equiv M/N-1$\footnote{Considering one
  $N$th of words as rare means that the average interval length is
  almost $N$, for any given total number of words $M$, as follows. One
  $N$th of words means $M/N$ words. Then, for sufficiently large $M$,
  the mean interval length is $(M-1) / (M/N-1) \approx N$.}. For the
resulting interval sequence $R_N=r_1,r_2,\ldots,r_{M_N}$, let the mean
and standard deviation be $\mu_N$ and $\sigma_N$, respectively. Then
the autocorrelation function is calculated for $R_N$ with $M_N$,
$\mu_N$, $\sigma_N$ replacing $M$, $\mu$, $\sigma$, respectively, in
formula (\ref{acf}).

For literary texts, $C(s)$ take positive values forming a power law
\citep{plosone16}.  The blue points in the rightmost graph in
\figref{fig:lesmiserables} represent the actual $C(s)$ values in a
log-log plot for a sequence of {\em Les Mis\'{e}rables} in its
entirety\footnote{The values of $s$ were taken up to $M_N$/100 in a
  logarithmic bin, following \citep{plosone16}, which is the limit for
  the resulting $C(s)$ values to remain reliable, based on the
  previous fundamental research such as reported in
  \citep{Lennartz2009}. For $s$ larger than $M_N$/100, the values
  of plots tend to decrease rapidly.  }. The thick gray line
  represents the fitted power-law function, which shows that this
  degree of clustering decays by a power law with exponent
  $\gamma=0.301$ and a fit error of 0.00158 per point. The fit error
  reported in this article is the average distance from the fitted
  line for a point, or in other words, the root of all the accumulated
  square errors, divided by the number of points. The points were all
  positive within the chosen range of $s$.

As mentioned previously, the analysis scheme explained above uses
extreme value analysis in addition to interval analysis. The method
was established within the statistical physics domain, originally for
analyzing extreme events with numerical data, such as devastating
earthquakes. Analysis schemes using intervals between such rare events
always consider rarer events above a threshold (corresponding here to
$N$), in order to tackle the low-frequency problem. Various complex
systems are known to exhibit long-range correlation (or long-range
memory), as reported in the natural sciences and finance, e.g.,
\citep{Corral1, C4, Bunde2005, Kantz,
  Blender2015,Turcotte,Yamasaki,Bogachev}. Rare words in a language
sequence should then correspond to extreme events, and the analysis
scheme was hence developed as reported in \citep{plosone16}. That work
showed how 10 single-author texts exhibit long-range
correlation. Thus, among multiple reports so far, there is abundant
evidence arguing that language is long-range correlated in the word
arrangement.

Therefore, following that previous work, long-range correlation is
reconsidered here through CHILDES data and mathematical generative
models. In \citep{plosone16}, $N$ was varied across
$2,4,8,16,32,64$. For large $N$, the interval sequence becomes too
short for proper analysis, but for small $N$, it includes words that
occur too frequently. To focus on the main point of the article
without having too many parameters, $N=16$ is used throughout the
remainder.

The main contribution of this paper is to discuss Bayesian models in
seeking the reason why such long-range correlation exists. Before
proceeding, two other, more common power laws are introduced because
they are necessary for the later discussion in \secref{sec:model}. The
leftmost graph in \figref{fig:lesmiserables} shows the log-log
rank-frequency distribution for {\em Les Mis\'{e}rables}, which
demonstrates a power-law relationship between the frequency rank and
frequency, i.e., Zipf's law.  Given word rank $u$ and frequency $F(u)$
for a word of rank $u$, Zipf's law suggests the following
proportionality formula:
\begin{equation}
F(u) \propto u^{-\xi}, \ \  \msk \xi \approx 1.0. \label{eq:zipf}
\end{equation}
As shown here for {\em Les Mis\'{e}rables}, the plot typically follows
formula (\ref{eq:zipf}) only very approximately. A skew or bias, such
as the convex tendency to the upper right, often appears, as will be
seen for the CHILDES corpus in the following section. There have been
discussions on how to improve the Zipf model by incorporating such
bias \citep{mandelbrot} \citep{mandelbrot_book} \citep{montemurro01}
\citep{deng14} \cite{altmanprx}. To the best of the author's knowledge, however, 
mathematical model fully explains the bias is still under debate.

The middle graph in \figref{fig:lesmiserables} shows the type-token
relation based on another power law, usually referred to as Heaps'
Law, indicating the growth rate of the vocabulary size with respect to
text length. Given vocabulary size $V(m)$ for a text of length $m$,
Heaps' law is as follows:
\begin{equation}
  V(m) \propto m^{\zeta}, \ \  \msk \zeta < 1.0. \label{eq:heaps}
\end{equation}
This feature was known even before \citep{heaps}, as published in
\citep{Herdan1964} and \citep{Guiraud1954}. In the graph, the black
line represents an exponent of 1.0. As seen here, for {\em Les
  Mis\'{e}rables}, $\zeta=0.672$, much smaller than 1.0; indeed, the
growth rate for natural language is below 1.0.

\section{Autocorrelation Functions for Childhood Language}
\label{sec:child}
Using the method introduced in the previous section, this section
introduces a kind of data, which has never been considered in
the context of long-range correlation: childhood language data from
the CHILDES corpus. In contrast to the previous work on single-author
texts, these data concern utterances (speech). Furthermore, the data
are chronologically ordered, thus showing the development of a child's
linguistic capability.

\begin{figure} [t]
\centering
\includegraphics[width=\textwidth]{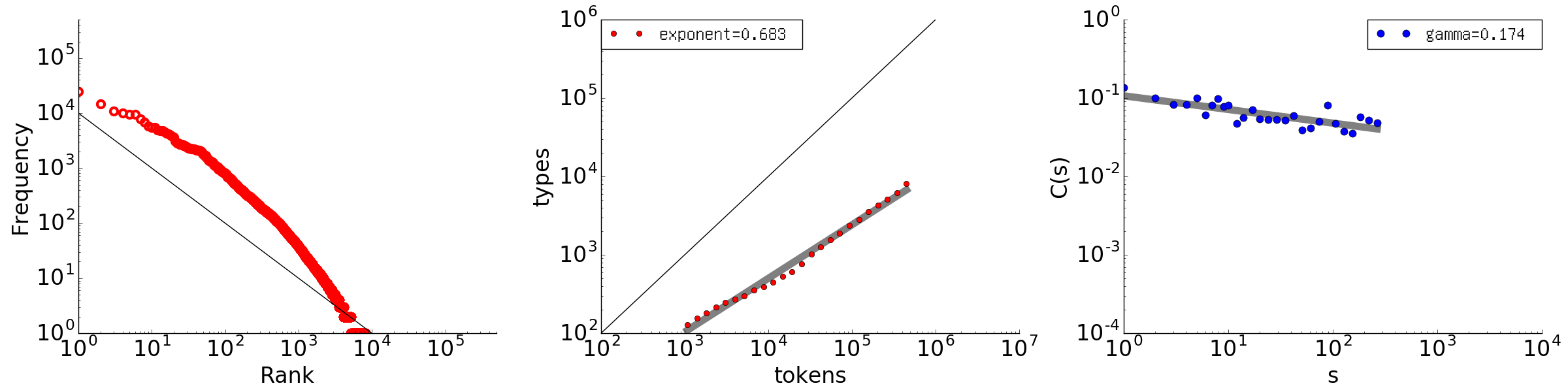}
\caption{Rank-frequency distribution, type-token relation, and
  autocorrelation function for the Thomas data set (103762 words).}
\label{fig:thomas}
\end{figure}

The first example is Thomas in English, which is the longest data set
in CHILDES.  \figref{fig:thomas} shows the rank-frequency
distribution, type-token relation, and autocorrelation function for
Thomas' utterances, similarly to \figref{fig:lesmiserables}.

The autocorrelation function (right) has a surprisingly tight power
law, thus indicating long-range correlation. Since a child's
utterances are linguistically under development, this result is not
trivial. The slope is smaller than that of the literary text shown in
\figref{fig:lesmiserables}. None of the calculated $C(s)$ values were
negative, and the fit error was 0.00255 per point.

As for the rank-frequency distribution (left), the overall slope is
larger than 1.0 and the plot has a clear convex tendency, as compared
with the black line representing a slope of 1.0. Such convex tendency
of the rank-frequency distribution has been reported elsewhere, such
as for single-author collections \citep{montemurro01} or Chinese
characters \citep{deng14}, but the convexity here seems different from
both of those cases. This convex tendency suggests, rather, that
Thomas generated utterances using more frequent words, especially the
top 100 words.

Lastly, the middle graph shows the type-token relation. As compared
with {\em Les Mis\'{e}rables}, the vocabulary growth is less stable
and slightly steeper, with an exponent of $\zeta=0.683$.

\begin{figure} [t]
\centering
\includegraphics[width=10cm]{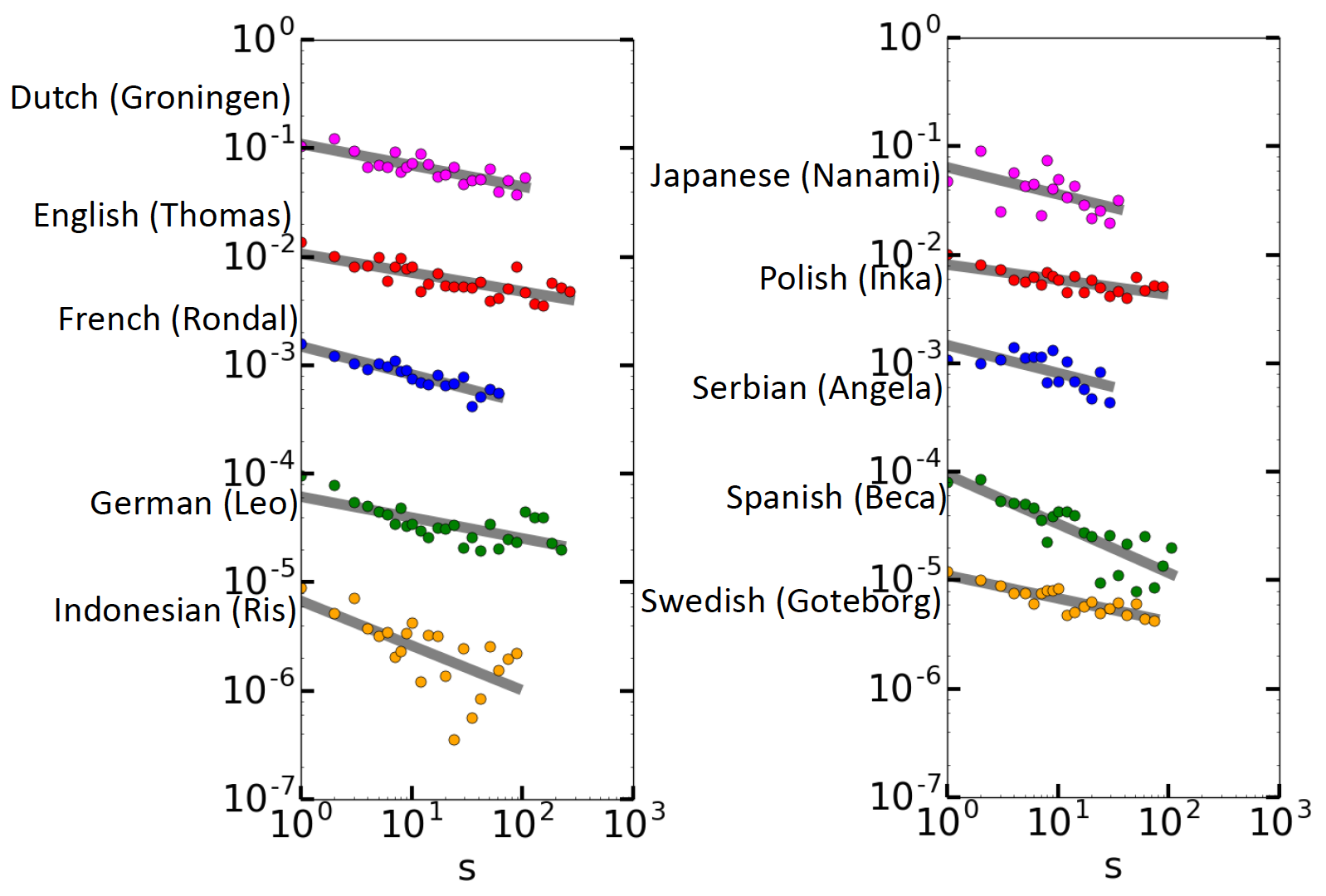}
\caption{Autocorrelation functions for the 10 children. For the sake
  of vertical placement, the $C(s)$ values for the $z$th data set from
  the top are multiplied by $1/10^{z-1}$ in each graph.}
\label{fig:acf10}
\end{figure}

The 10 longest CHILDES data sets were selected, and these included
utterances in different languages. The utterances were carefully
separated by speaker, and only those by children were used. Moreover,
the CHILDES codes for unknown words were removed.
\figref{fig:acf10} shows the autocorrelation function results for
Thomas and the other nine children. Although not always as tight as in
Thomas's case, the power law does hold in every case. Except for a
single point at $s=10$ for Ris, all calculated $C(s)$ values are
positive and aligned almost linearly. Moreover, the power law holds
more tightly for the larger data sets.

\section{Generative Language Models}
\label{sec:model}

\begin{figure} [t]
\centering
\includegraphics[width=\textwidth]{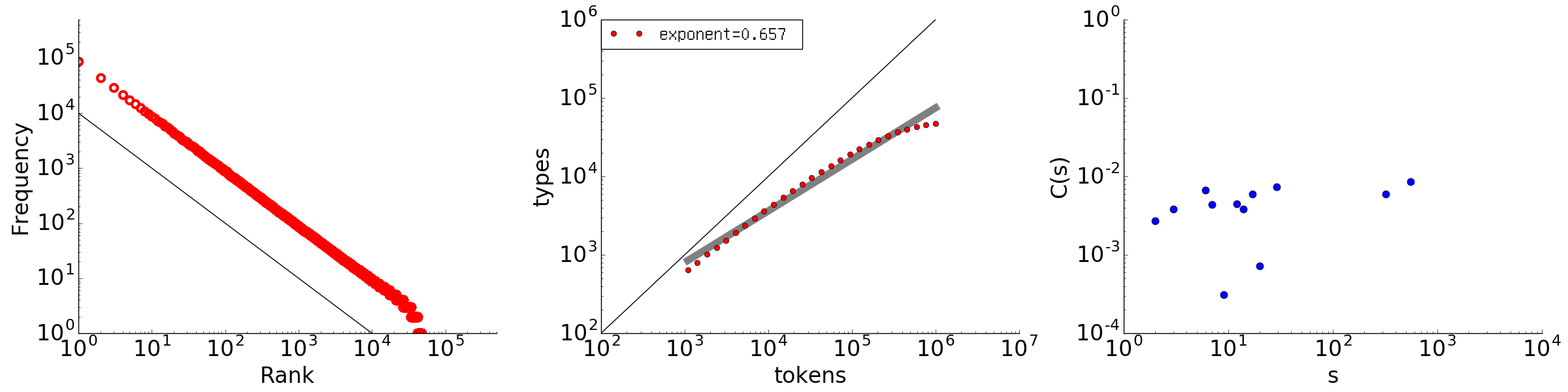}
\caption{Rank-frequency distribution, type-token relation, and
  autocorrelation function for a randomly generated sequence that
  follows a Zipf distribution (one million words with a vocabulary
  size of 50000).}
\label{fig:zipf}
\end{figure}

The autocorrelative characteristic reported here for children's
utterances and in many previous works for natural language texts does
not hold for simple random data. To demonstrate this, three examples
are provided. The first example is a randomized word set whose
rank-frequency sequence strictly follows a Zipf
distribution\footnote{The random sequence was generated by an inverse
function method.}. \figref{fig:zipf} shows graphs of the
rank-frequency distribution, type-token relation, and autocorrelation
function for this sequence, with a length of one million words and a
vocabulary size of 50000 words.

The leftmost graph does exhibit a power law with the exponent $-1.0$,
but the rightmost graph shows that the long-range correlation is
completely destroyed. Many $C(s)$ values are negative and thus not
shown here because the plot is log-log. As noted before, for random
data the autocorrelation function fluctuates around 0. Approximately
half the values become negative and thus disappear from the figure,
leaving a sparse set of plotted points, exactly as observed here.

\begin{figure} [t]
\centering
\includegraphics[width=\textwidth]{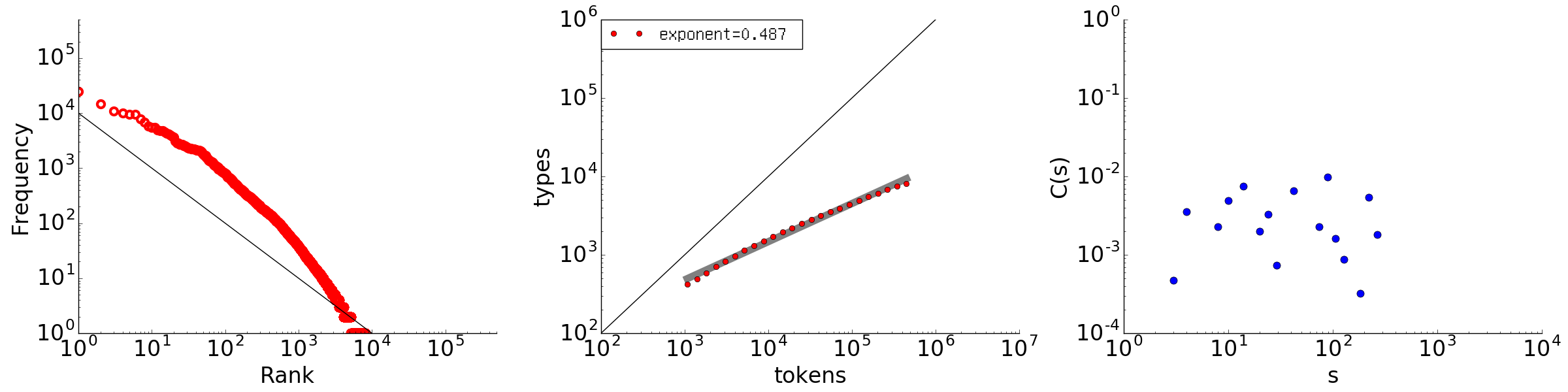}
\caption{Rank-frequency distribution, type-token relation, and
  autocorrelation function for the Thomas data set randomly shuffled
  at the word level.}
\label{fig:shuffle_thomas}
\end{figure}

A second example was obtained by shuffling Thomas's utterances at the
word level.  Random shuffling destroys the original intervals between
words in the Thomas data set.  \figref{fig:shuffle_thomas} shows the
analysis results, in which the autocorrelation function has become
random, whereas the rank-frequency distribution and type-token
relation remain the same as the original results shown in
\figref{fig:thomas}.

\begin{figure} [t]
\centering
\includegraphics[width=\textwidth]{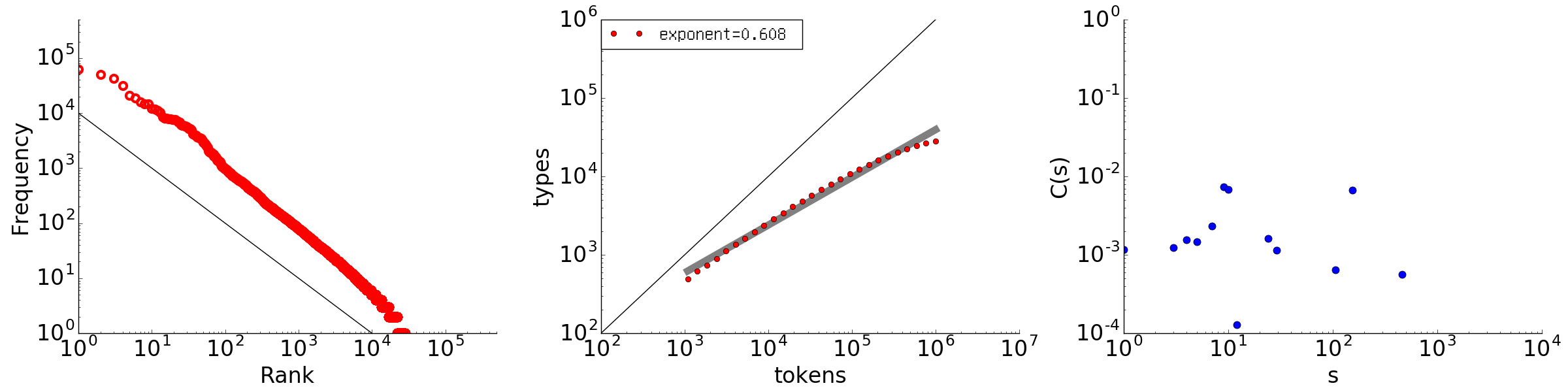}
\caption{Rank-frequency distribution, type-token relation, and
  autocorrelation function for a sequence randomly generated from
  bigram (1st-order Markov) models of {\em Les Mis\'{e}rables} (one
  million words).}
\label{fig:bigram_lesmis}
\end{figure}

The third example is a Markov sequence generated using bigrams
obtained from {\em Les Mis\'{e}rables}. The random sequence was
generated from the bigrams according to the probabilities recorded in
a word transition matrix. \figref{fig:bigram_lesmis} shows the
analysis results, with the rightmost graph indicating that the
autocorrelation function does not exhibit any memory.

Long-range correlation therefore does not hold for such simple random
sequences. At the same time, given that long-range memory holds for
the CHILDES data, it should be natural to consider that some simple
mechanism underlies language production. In early childhood speech,
utterances are still lacking in full vocabulary, ungrammatical, and
full of mistakes. Therefore, the long-range correlation of such speech
must be based on a simple mechanism other than linguistic features
such as grammar that we generally consider.

The problem with all the findings related to power laws underlying the
statistical physics domain is that even though, as mentioned before,
the method has been effective in analyzing natural sciences and
finance, the exact reason why such power laws hold remains
unknown. This applies to long-range correlation, as well: ``rare words
tend to cluster'' is only one simplistic way to express a limited
aspect of the phenomenon. As mentioned before, however, the phenomenon
is more complex and has some relation to the scale-free property
underlying language.

In the case of Zipf's law, Mandelbrot mathematically proved that it
holds by optimizing the communication efficiency
\citep{mandelbrot}\citep{mandelbrot_book}. It is unknown how this
optimization theory could relate to long-range correlation. Moreover,
it is not obvious whether an infant child would optimize every word of
an utterance. It would be more natural to consider that a child learns
how to act in choosing a word, and that this action, in fact, is
mathematically bound so as to be optimal. One possible approach to
understand what's behind such an {\em act} would be to consider the
behavior of mathematical models of language with respect to power
laws.  Roughly, at least four representative families of mathematical
processes have been considered as language models: Markov models,
Poisson processes \cite{church} or renewal processes
\cite{Altmann2009}, neural languages models, and recent Bayesian
models. The first two models require a predefined vocabulary size, so
without further modification of these models, they could not be
applied to confirm either Zipf's or Heaps' law.  Neural language
models have been successful in language applications, but has been
reported recently that they do not exhibit long range
correlation \cite{plos18}.

The rest of this paper therefore focuses on Bayesian models, which
naturally accommodate infinite vocabulary growth. In all the
generative models presented hereafter, the model generates elements
one after another, either by introducing a new word or by reusing a
previous element. Let $K_t$ be the number of kinds of elements
(vocabulary size) at time $t$, and let $S_{t,i}$ be the frequency of
elements of kind $i$ occurring until $t$. At $t=0$, all models
presented hereafter start with the following status:
\[
  K_0 = 1,\msk S_{0,1} = 1, \msk S_{0,i} = 0, \msk i \in \mathbb{Z}_{> 1}.
\]

The most fundamental model is the Simon model \citep{simon55}
\citep{mitzenmacher03}. This model, described colloquially as ``the
rich get richer,'' is used for a variety of natural and artificial
phenomena. A similar model in complex network systems is the
Barab\'{a}si-Albert model \citep{ba99}. For $t > 0$, given a constant
$ 0 < \alpha < 1 $, an element is generated at time $t+1$ with the
following probabilities:
\begin{eqnarray*}
\displaystyle
P(K_{t+1} = K_t+1, S_{t+1,j} = S_{t,j}, j\in \mathbb{Z}_{\geq 1}\backslash\{K_t+1\}, S_{t+1, 
K_t+1} = 1)
&=& \alpha, \\
P(K_{t+1} = K_t, S_{t+1,i} = S_{t,i} + 1, S_{t+1,j} = S_{t,j}, j \in \mathbb{Z}_{\geq 
1}\backslash\{i\})
&=& (1-\alpha) \frac{S_{t,i}}{t}, i = 1,\ \ \ldots,K_t.  
\end{eqnarray*}
Note that the first definition gives the case when a new word is
introduced, and the second gives the case when a previous word is
sampled. The scheme can thus be described as follows: with constant
probability $\alpha$, a new, unseen element is generated; and with the
remaining probability $1-\alpha$, an element that has already occurred
is selected according to the frequency distribution in the
past. Suppose, for example, that the previously generated sequence is
$X$=$[$`x', `y', `x', `z', `x', `z'$]$.  Then the next element will be
a new element with probability $\alpha$, or `x', `y', or `z' with
probability $3(1-\alpha)/6$, $(1-\alpha)/6$, or $2(1-\alpha)/6)$,
respectively. It is trivial to understand that this reuse of previous
elements is equivalent to a {\em uniform sample} from the past
sequence, i.e., by considering that all past elements occurred equally
under a uniform distribution. In this example, uniform sampling
entails picking one element randomly from $X$=$[$`x', `y', `x', `z',
  `x', `z'$]$.

It has been mathematically proven that the rank-frequency distribution
of a sequence generated with the Simon model roughly follows a power
law, independently of the value of $\alpha$
\citep{mitzenmacher03}. Since the vocabulary introduction rate is
constant, it is trivial to see that the type-token ratio also has the
exponent $1.0$.

\begin{figure} [t]
\centering
\includegraphics[width=\textwidth]{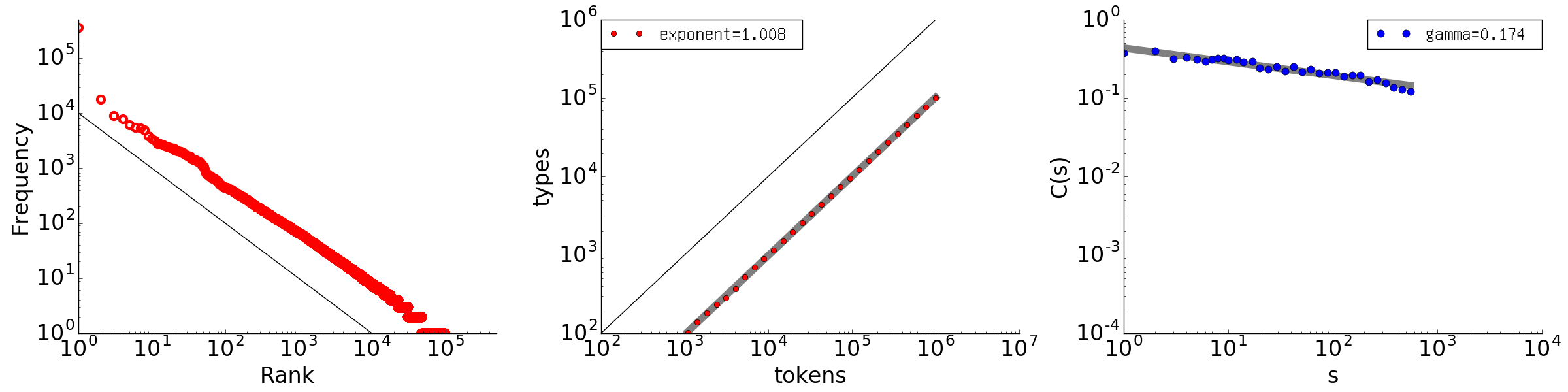}
\caption{Rank-frequency distribution, type-token relation, and
  autocorrelation function for a one-million-element sequence
  generated with the Simon model, where $\alpha = 0.10$.}
\label{fig:simon}
\end{figure}

To investigate the Simon model, a sequence of one million elements
with $\alpha = 0.10$ was generated, and its rank-frequency
distribution, type-token relation, and autocorrelation function were
obtained. The autocorrelation function was calculated according to the
scheme explained in \secref{sec:acf}, since a new element introduced
in this scheme can be anything, even a non-numerical element.

\figref{fig:simon} shows the results. The first two graphs agree with
the theory by giving exponents of $-1.0$ and $1.0$, respectively. As
for the autocorrelation function, surprisingly, long-range memory is
clearly present. The slope is $\gamma=0.174$, which is coincidentally
the same as that for the Thomas data set. None of the $C(s)$ values is
negative, and the fit to the slope is very tight, with a fit error of
0.00369.

To examine the parameter dependence, 10 sequences for each of
$\alpha={0.1,0.2,0.3,0.4}$ were generated, and the autocorrelation
function was obtained for each. The results included no negative
$C(s)$ values. For each $\alpha$, the respective mean values of
$\gamma$ were 0.156, 0.133, 0.118, and 0.095, with small standard
deviations of 0.019, 0.018, 0.011, and 0.013, respectively. Thus, the
slope decreased with increasing $\alpha$. The average fit error
obtained via the square error across all 40 sequences was 0.00366 per
point.

The Simon model has two known problems, however, as a language
model. The first is that the vocabulary growth (proven to have
exponent 1.0) is too fast. Indeed, such fast vocabulary growth is very
unlikely in natural language production. The second problem is that
the model cannot handle the convexity underlying a rank-frequency
distribution, as observed especially for the Thomas data set
(\figref{fig:thomas}). Such convexity has been reported elsewhere, as
noted before.

Another Bayesian model called the Pitman-Yor model \citep{pitman}
solves these two problems. Using the same mathematical notation as
before, and given two constants $0 \leq a < 1$ and $0 \leq b$, the
following generative process is applied for $t > 0$ at time $t+1$ with
the following probabilities:
\begin{eqnarray*}
P(K_{t+1} = K_t+1, S_{t+1,j} = S_{t,j}, j\in \mathbb{Z}_{\geq 1}\backslash\{K_t+1\}, S_{t+1, 
K_t+1} = 1)
&=&  \frac{a K_t + b}{t + b}, \\
P(K_{t+1} = K_t, S_{t+1,i} = S_{t,i} + 1, S_{t+1,j} = S_{t,j}, j \in \mathbb{Z}_{\geq 
1}\backslash\{i\})
&=& \frac{S_{t,i} - a}{t+b}, i = 1,\ \  \ldots,K_t.  
\end{eqnarray*}
As with the Simon model, the first line defines the introduction rate
for new elements. It decreases with the length of the sequence, $t$,
yet is linear in the vocabulary size $K_t$ according to the strength
$a$. This amount is generated as a sum of taking every element kind
$i=1,\ldots,K_t$ by subtracting $a$ from frequency $S_{t,i}$, which
appears in the numerator of the second definition above. Apart from
this, the parameter $b$ controls the convex trend \citep{pitman}
\citep{teh06} often seen in rank-frequency distributions. When $a=0$,
this model reduces to the Chinese restaurant process
\citep{goldwater}, which has been applied widely in the language
engineering domain.

Mathematically, the parameter $a$ in the Pitman-Yor model almost
equals the value of the exponent of the type-token relation, $\zeta$,
which describes the vocabulary growth speed, provided that $b$ is
small and Heaps' law holds (Appendix A). According to empirical
verification, even for a large $b=10000$, $\zeta$ only differed from
$a$ by a maximum of 0.1. Given this, $a=0.68$ was chosen for the
remaining Pitman-Yor models presented in this article, a value
somewhat in the middle of $\zeta=0.683$ for the Thomas data set and
$\zeta=0.672$ for {\em Les Miserables}.

\begin{figure} [t]
\centering
\includegraphics[width=\textwidth]{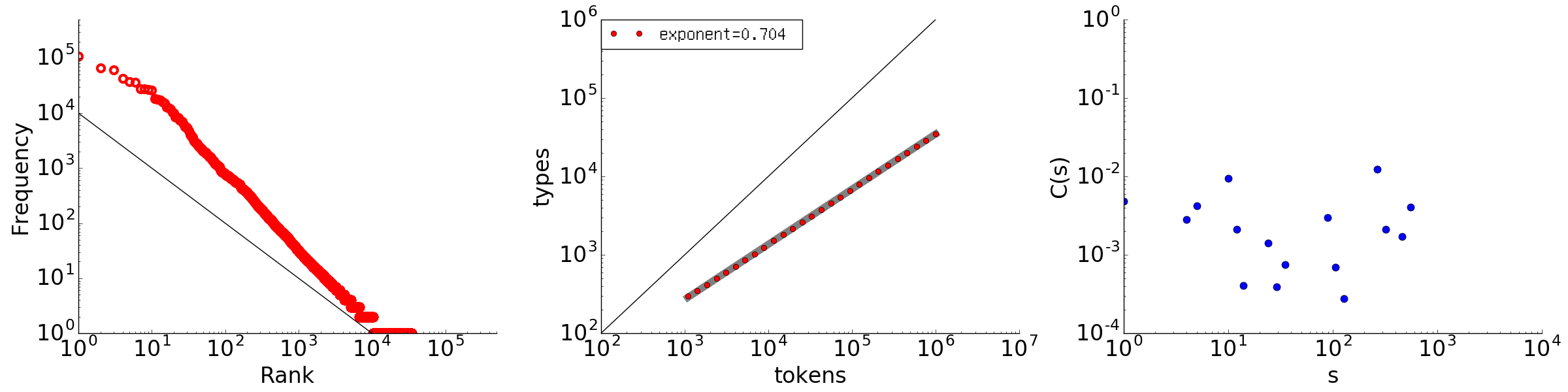}
\caption{Rank-frequency distribution, type-token relation, and
  autocorrelation for a sequence of one million elements generated by
  a Pitman-Yor model with $a=0.68$ and $b=0.80$.}
\label{fig:pitmanyor}
\end{figure}

For generation of one million elements by a Pitman-Yor process with
$a=0.68$ and $b=0.80$, \figref{fig:pitmanyor} shows the three
resulting graphs. Agreeing with theory, the middle graph showing the
type-token relation has a slope reasonably close to 0.68. As for the
leftmost graph showing the rank-frequency distribution, with $b=0.80$
a slight convex tendency appears, but with a larger $b \geq 100$, the
convex tendency would clearly be present. In the rightmost figure,
however, the power law of particular interest here, for the
autocorrelation function, has disappeared. Although the change from
the Simon model is subtle, with respect to the value of $a$, the
sequence does not exhibit any arrangement underlying natural language.

Since this result could be due to the parameter setting, all possible
combinations of $a=\{0.0,0.1,0.2,\dots,0.9\}$ (10 values) and
$b=\{0.0,0.1,0.2,\ldots,1.0,10.0,100.0,1000.0,10000.0\}$ (15 values)
were considered.  For every pair $(a,b)$ out of these 150
possibilities, a sequence of one million elements was generated and
examined for long-range correlation. If any $C(s)$ value for $s < 10$
was negative, then long-range memory was judged not to hold. This
criterion is somewhat loose, because it considers long-range
correlation to hold even when the points are scattered and not
exhibiting power-law behavior, as long as they are still
positive. Even with this loose criterion, however, none of the
sequences had long-range correlation. When $a$ is too small, the rate
of introducing new words becomes too weak.  Even when there are
sufficient new words, the arrangement seems qualitatively different
from the case of the Simon model.

\begin{figure} [t]
\centering
\includegraphics[width=\textwidth]{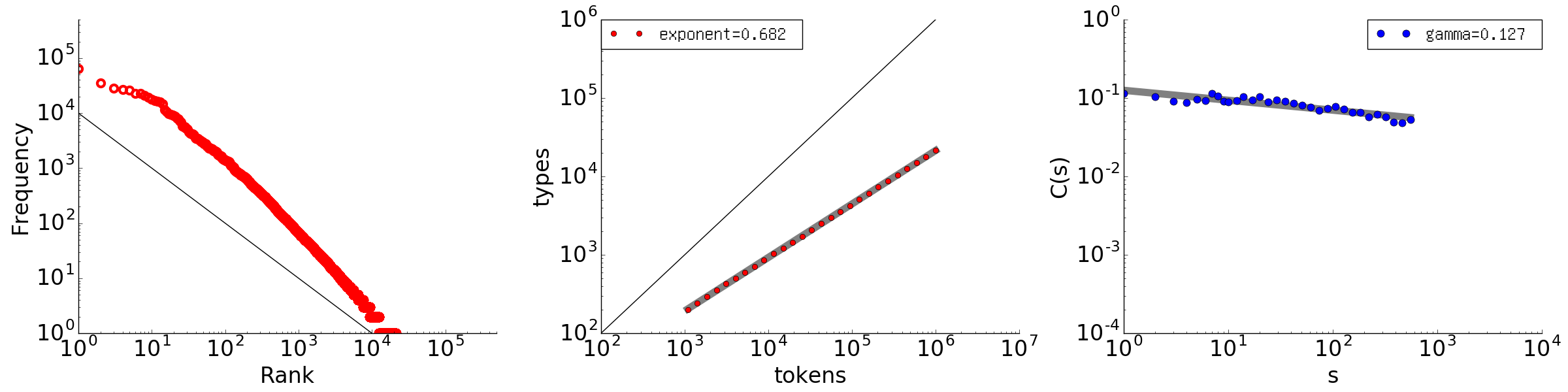}
\caption{Rank-frequency distribution, type-token relation, and
  autocorrelation for a sequence of one million elements generated by
  the proposed conjunct model with $a=0.68$ and $b=0.80$.}
\label{fig:decr}
\end{figure}

We have now seen that the Simon model exhibits a bad type-token
relation but a good autocorrelation, while the opposite is true for
the Pitman-Yor model. Since long-range correlation is due to the
arrangement of frequent words and rare words, a natural approach is to
test the following conjunct generative model for $t > 0$ at time $t+1$
with the following probabilities:
\begin{eqnarray*}
P(K_{t+1} = K_t+1, S_{t+1,j} = S_{t,j}, j\in \mathbb{Z}_{\geq 1}\backslash\{K_t+1\}, S_{t+1, 
K_t+1} = 1)
&=&  \eta, \ \ \ where\ \  \eta = \frac{a K_t + b}{t + b},\\
P(K_{t+1} = K_t, S_{t+1,i} = S_{t,i} + 1, S_{t+1,j} = S_{t,j}, j \in \mathbb{Z}_{\geq 
1}\backslash\{i\})
&=& (1-\eta) \frac{S_{t,i}}{t}, i = 1, \ \ \ldots,K_t.  
\end{eqnarray*}
This mixed model introduces new words with a probability $\eta$ equal
to that of the Pitman-Yor model, so the first line is exactly the same
as in the definition of that model.  As for sampling, with probability
$1-\eta$ a previous element is introduced in proportion to the
frequencies of the elements. In other words, the conjunct model
achieves uniform sampling, as in the Simon model, by replacing that
model's $\alpha$ with $\eta$.

\begin{figure} [t]
\centering
\includegraphics[width=0.6\textwidth]{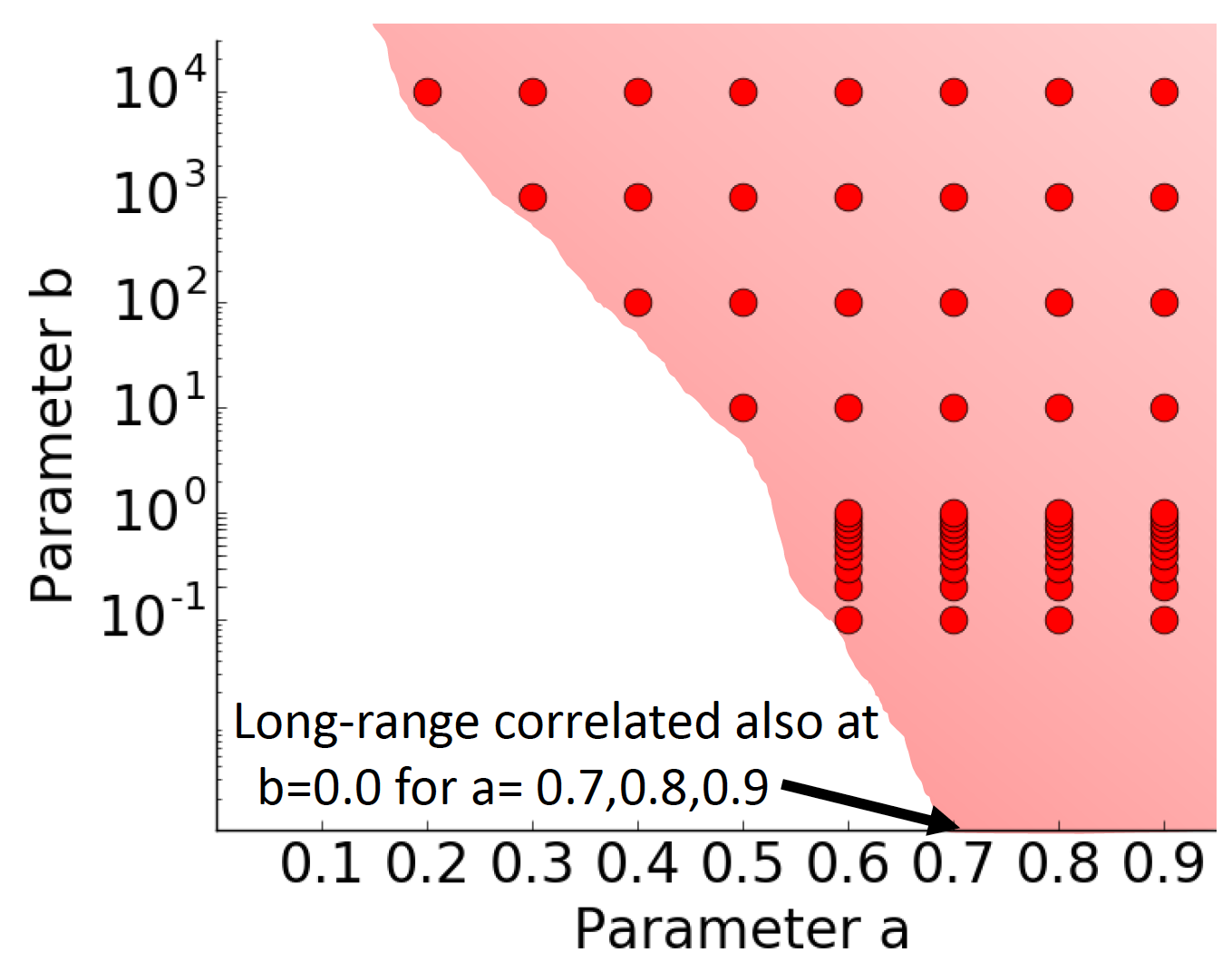}
\caption{Pairs of parameters $a$ and $b$ for which a sequence
  generated by the conjunct model exhibits long-range correlation. The
  red dots represent the experimental results, while the interpolated
  area of long-range correlation was manually shaded red.}
\label{fig:parameter}
\end{figure}

\figref{fig:decr} shows the behavior of a sequence generated by the
conjunct model with $a=0.68$ and $b=0.80$. The model clearly exhibits
the desired vocabulary growth while maintaining its long-range
correlation. The $\gamma$ exponent decreases to 0.127, with a fit
error of 0.00162.

To examine the parameter dependence, again all possible combinations
of 10 values of $a=\{0.0, 0.1,0.2,\dots,0.9\}$ with
$b=\{0.0,0.1,0.2,\ldots,0.9,1.0,10.0,100.0,1000.0,10000.0\}$ (15
values) were again considered. For every pair $(a,b)$, a sequence of
one million elements was generated 10 times and examined for
long-range correlation. \figref{fig:parameter} shows all the pairs of
values for which long-range correlation was observed. When $a$ is too
small, the rate of introducing new words becomes too small and gives
no correlation. For a sufficiently large $a$ and a value of $b$ that
depends on $a$, on the other hand, long-range correlation is observed.

The variance of the obtained exponent $\gamma$ for the autocorrelation
function was almost the same for small $b$ below 1.0. For example, for
$a=0.68$ the average $\gamma$ was 0.126 with a standard deviation of
0.0318 and fit error per point of 0.0013. For larger $b$, $\gamma$
tended to be smaller. For other $a$ values, as well, the $\gamma$
values were not as steep as 0.2.

\section{Discussion}

The findings reported in this article lead to two main points. First,
the findings raise the question of the Pitman-Yor model's validity as
a language model. Pitman-Yor models have been used because they nicely
model the rank-frequency distribution and the growth rate of natural
language. However, 
the Pitman-Yor model
is not long range correlated, different from natural language. 
Although the current work does not invalidate the
usefulness of Pitman-Yor models for language engineering (since they
are effective), the long-range correlation behavior does reveal a
difference between the nature of language and a Pitman-Yor model. This
could be a factor for consideration in future scientific research on
language.

Second, this work reveals that among possible mathematical language
models considered so far, those with uniform sampling generated
strong long-range correlation (i.e., the Simon model and the conjunct
model developed at the end of the previous section).  Given how simple
uniform sampling is, however, the findings could suggest that natural
language has some connection with uniform sampling.  In fact,
long-range correlation is present in music, as well (Appendix B),
which is another human resource, similar to language.  Therefore,
the human faculty to generate linguistic-related time series might
have a fundamental structure with some relation to a very simple
procedure, with uniform sampling as one possibility.

Without noting, however, uniform sampling only by itself is limited as
a language model. In addition to the lack of linguistic grammatical
features, the Simon model and its extensions exhibit different nature
at the beginning part and later part of the sample: this is different
from language, where a sample from any location of the data is long
range correlated.  Mathematical generative processes that satisfy all
the stylized facts of languages would aid to clarify what kind of
process language is, and to this end, the proposed conjunct model
could be yet another starting point towards a better language model.
The conjunct model currently has two differences from actual natural
language. The first is the exponent $\gamma$, which is larger for both
literature and the CHILDES data, sometimes exceeding 0.3, but remains
below around 0.15 for the conjunct model. Second, the rank-frequency
distribution is convex for large $b$ in the conjunct model, but such
large $b$ makes $\gamma$ even smaller. Therefore, the conjunct model
must be modified to fill this gap.  This would require more exhaustive
knowledge of long-range memory in natural language, and the model
would have to integrate more complex schemes that possibly introduce
n-grams or grammar models.

\section{Conclusion}
This article has investigated the long-range correlation underlying
the autocorrelation function with CHILDES data and Bayesian models by
using an analysis method for non-numerical time series, which was
borrowed from the statistical physics domain.  After first overviewing
how long-range correlation phenomena have been reported for different
kinds of natural language texts, they were also verified to occur for
children's utterances.

To find a reason for this shared feature, we investigated three
generative models: the Simon model, the Pitman-Yor model, and a
conjunct model integrating both. The three models share a common
scheme of introducing a new element with some probability and
otherwise sampling from the previous elements. The Simon model
exhibits outstanding long-range correlation, but it deviates from
natural language texts by causing the vocabulary to grow too fast. In
contrast, the Pitman-Yor model exhibits no long-range correlation,
despite having an appropriate vocabulary growth rate. Therefore, the
conjunct model uses the Pitman-Yor introduction rate for new
vocabulary but samples from the past through uniform sampling, like
the Simon model. This conjunct model produces long-range correlation
while maintaining a growth rate similar to that of natural language
text.

The fact that the Pitman-Yor model does not exhibit long-range
correlation raises the question of the Pitman-Yor model's validity as
a natural language model.  Since the mathematical generative models
among Simon kind that exhibited long-range correlation are based on
uniform sampling, we may conjecture that there could be some relation
between natural language and uniform sampling. The findings in this
article could provide another direction towards better future language
models.

\section*{Appendix A}
This appendix explains why $a \approx \zeta$. At time $t$, the number
of words introduced into the sequence is
\begin{equation}
\displaystyle
  K_t = \int_0^t \frac{a K_t + b}{t + b}.
\end{equation}
Assuming that $b$ is sufficiently small and $K_t=t^\zeta$, 
\begin{eqnarray}
\displaystyle
\int_0^t \frac{a t^\zeta + b}{t + b} &\approx& \int_0^t a t^{\zeta-1}  \\
&=& \frac{a}{\zeta} t^{\zeta}, 
\end{eqnarray}
and since this must equal $K_t$, $a=\zeta$. Empirically, this was almost always the 
case for $b$ up to around 1.0. For larger $b$, $\zeta$ became larger than $a$: when 
$b=10000$, for example, $\zeta$ was larger than $a$ by 1.0, at most. 

\section*{Appendix B}

\begin{figure} [t]
\centering
\includegraphics[width=10cm]{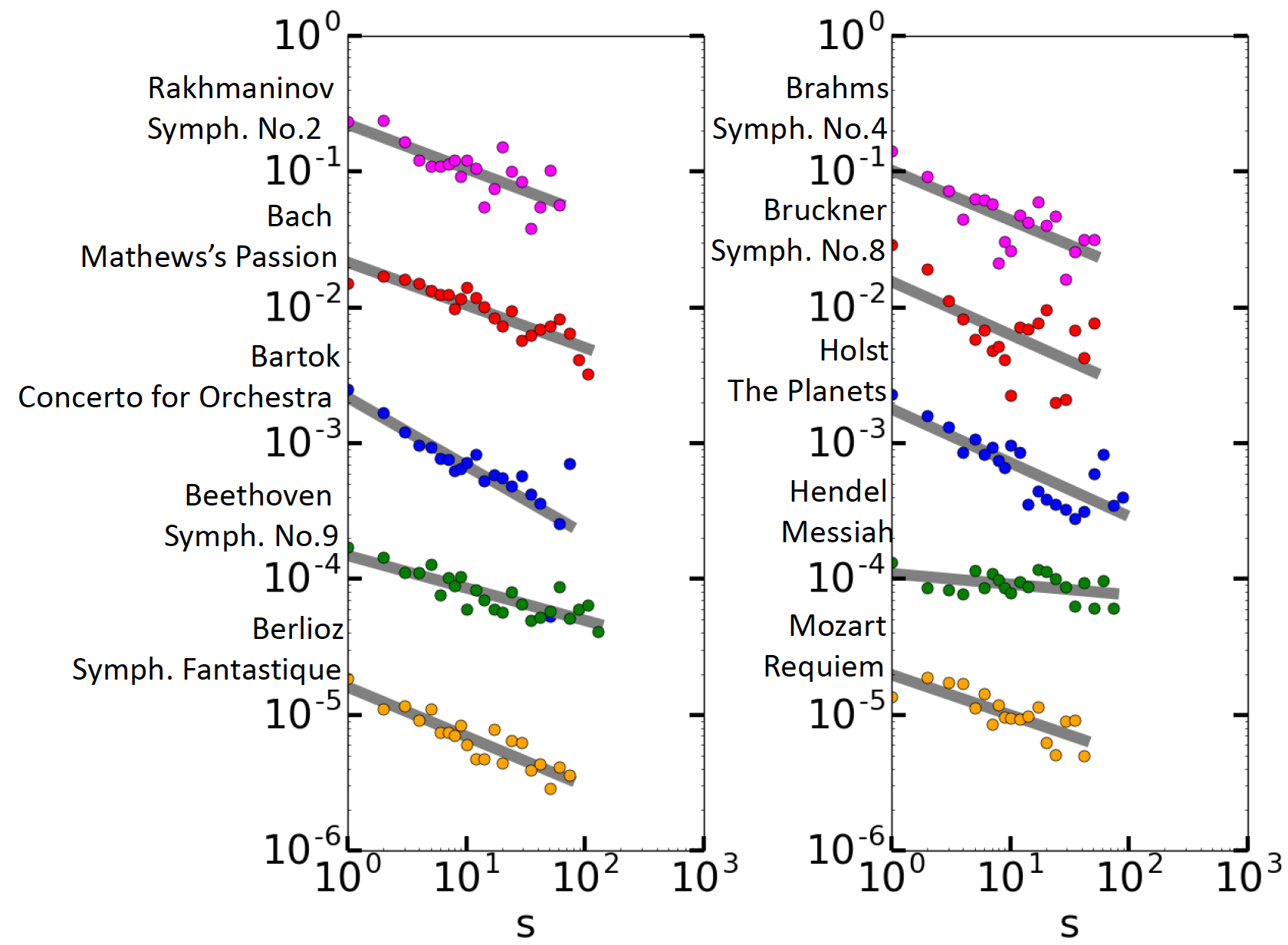}
\caption{Long range correlation of 10 long classical music pieces.
  For the sake of vertical placement, the C(s) values for the zth data
  set from the top are multiplied by $1/10^{z-1}$ in each graph.  
\label{fig:music}}
\end{figure}

This Appendix shows the long range correlation of 10 long classical
music pieces (\figref{fig:music}). 
Original data was in MIDI format and they are
pre-processed.  The headers and footers were eliminated and so that the
data contain only the musical part, including pause indications.
Every tune played by different instrument kind is separated and
concatenated As far as these 10 pieces, the long range correlation can
be considered as to hold.  Previous work on also reports how
long-range correlation holds in skilled piano play\cite{ruiz2014}.

\section*{Acknowledgement}
I would like to thank the PRESTO program, of the Japan Science and
Technology Agency, for its financial support. \\

\bibliographystyle{natbib}
\bibliography{gm}

\end{document}